\PassOptionsToPackage{table}{xcolor}
\PassOptionsToPackage{dvipsnames}{xcolor}
\documentclass{article}

\usepackage[preprint]{neurips_2025}
\usepackage{svg}
\usepackage{transparent} 

\usepackage[utf8]{inputenc} 
\usepackage[T1]{fontenc}    
\usepackage{hyperref}       
\usepackage{url}            
\usepackage{booktabs}       
\usepackage{amsfonts}       
\usepackage{nicefrac}       
\usepackage{microtype}      
\usepackage{xcolor}         
\usepackage{tcolorbox}
\usepackage{caption}
\usepackage{subcaption}
\usepackage{graphicx}
\usepackage{amsmath}
\usepackage{threeparttable}
\usepackage{makecell}
\usepackage{enumitem}
\usepackage{multirow}
\usepackage{longtable}
\usepackage{ulem}
\usepackage{color}
\usepackage{listings}
\usepackage{multirow}
\usepackage{float}
\usepackage{tabularx}
\usepackage{setspace}
\usepackage{marvosym}
\usepackage{algorithm}
\usepackage{algorithmic}
\usepackage{natbib}
\usepackage{marvosym}
\usepackage{tcolorbox}
\usepackage{authblk}
\setcitestyle{numbers,square}
\setcitestyle{open={[},close={]},citesep={,\!},numbers}

\title{Infi-MMR: Curriculum-based Unlocking Multimodal Reasoning via Phased Reinforcement Learning in Multimodal Small Language Models}

\author[1,†]{Zeyu Liu}
\author[2,†]{Yuhang Liu}
\author[3]{Guanghao Zhu}
\author[4]{Congkai Xie}
\author[1,4]{Zhen Li}
\author[5]{Jianbo Yuan}
\author[5]{Xinyao Wang}
\author[1]{Qing Li}
\author[6]{Shing-Chi Cheung}
\author[2]{Shengyu Zhang}
\author[2]{Fei Wu}
\author[1,4,\thanks{Corresponding to: \texttt{hongxia.yang@polyu.edu.hk}}]{Hongxia Yang}

\affil[1]{The Hong Kong Polytechnic University} 
\affil[2]{Zhejiang University}
\affil[3]{University of Electronic Science and Technology of China}
\affil[4]{Reallm Labs}
\affil[5]{Amazon}
\affil[6]{The Hong Kong University of Science and Technology}

\definecolor{lightgold}{rgb}{1.0, 0.95, 0.8}

\begin{document}

\maketitle

\begin{abstract}
  
  Recent advancements in large language models (LLMs) have demonstrated substantial progress in reasoning capabilities, such as DeepSeek-R1 \cite{guo2025deepseek}, which leverages rule-based reinforcement learning to enhance logical reasoning significantly. However, extending these achievements to multimodal large language models (MLLMs) presents critical challenges, which are frequently more pronounced for Multimodal Small Language Models (MSLMs) given their typically weaker foundational reasoning abilities: (1) the scarcity of high-quality multimodal reasoning datasets, (2) the degradation of reasoning capabilities due to the integration of visual processing, and (3) the risk that direct application of reinforcement learning may produce complex yet incorrect reasoning processes. To address these challenges, we design a novel framework \textbf{Infi-MMR} to systematically unlock the reasoning potential of MSLMs through a curriculum of three carefully structured phases and propose our multimodal reasoning model \textbf{Infi-MMR-3B}. The first phase, \textbf{Foundational Reasoning Activation}, leverages high-quality textual reasoning datasets to activate and strengthen the model's logical reasoning capabilities. The second phase, \textbf{Cross-Modal Reasoning Adaptation}, utilizes caption-augmented multimodal data to facilitate the progressive transfer of reasoning skills to multimodal contexts. The third phase, \textbf{Multimodal Reasoning Enhancement}, employs curated, caption-free multimodal data to mitigate linguistic biases and promote robust cross-modal reasoning. \textbf{Infi-MMR-3B} achieves both state-of-the-art multimodal math reasoning ability ($43.68\%$ on MathVerse testmini, $27.04\%$ on MathVision test, and $21.33\%$ on OlympiadBench) and general reasoning ability ($67.2\%$ on MathVista testmini). Resources are available at  \href{https://huggingface.co/Reallm-Labs/Infi-MMR-3B}{Infi-MMR-3B}.
\end{abstract}

\section{Introduction}\label{sec:introduction}
In recent years, large language models (LLMs)~\cite{guo2025deepseek,yang2024qwen2,jaech2024openai} have made remarkable strides in processing and generating human-like text across a wide range of domains. Conventional LLMs often rely on direct prediction to produce final outputs, typically overlooking the intermediate reasoning processes, which results in suboptimal performance on complex tasks. To address this limitation and meet the sophisticated demands of real-world applications, researchers are focused on enhancing the reasoning capabilities of LLMs. 

Notably, OpenAI~\cite{jaech2024openai} has leveraged complex Chain-of-Thought (CoT) datasets to train an LLM that demonstrates significant improvements in reasoning performance compared to its predecessors. Building on this approach, subsequent research has utilized high-quality, generated CoT reasoning data to further advance the reasoning proficiency of these models~\cite{lai2024step,lightman2023let,yao2023tree}. 
DeepSeek-R1~\cite{guo2025deepseek} introduces a highly efficient approach leveraging rule-based reinforcement learning, which substantially reduces the reliance on human-annotated reasoning data while enabling models to autonomously enhance their reasoning capabilities through exploration of question-answer pairs.

Despite these advancements, extending such achievements to multimodal large language models (MLLMs) poses significant challenges, particularly for models with limited parameters, such as those with 3B parameters, commonly referred to as Multimodal Small Language Models (MSLMs). They must efficiently integrate visual information with logical reasoning, a process that requires robust cross-modal processing and reasoning capabilities. Overall, this integration is hindered by three primary obstacles:
(1) \textbf{Lack of High-Quality Multimodal Reasoning Data}: Rule-based reinforcement learning (RL) demands verifiable answers, yet most multimodal tasks focus on captioning, image description, and visual question answering (VQA). Moreover, existing multimodal reasoning datasets predominantly address simple tasks, such as counting, with few providing both complex reasoning problems and verifiable answers.
(2) \textbf{Degradation of Basic Reasoning Capabilities in MSLMs}: The integration of visual and textual data in MSLMs often undermines their foundational reasoning abilities. Additionally, the complexity of cross-modal fusion can disrupt structured inference, leading to diminished performance on reasoning tasks~\cite{cheng2025comt,xiang2024atomthink}.
(3) \textbf{Complex but Unreliable Reasoning Steps}: Directly training MLLMs with RL to generate complex inference processes frequently results in protracted and inaccurate reasoning steps~\cite{huang2025vision}.

To address the above challenges and enhance the reasoning capability in MSLMs, we propose \textbf{Infi-MMR}, a curriculum-based progressive rule-based RL training framework that unfolds in three distinct phases:
\begin{itemize}[leftmargin=*, nosep]
\item \textbf{Foundational Reasoning Activation:} This phase focuses on developing reasoning capabilities from textual datasets. Rather than directly incorporating multimodal data, it exclusively utilizes high-quality textual reasoning data to strengthen the model's foundational reasoning abilities through reinforcement learning. This approach primes the model for robust logical reasoning, addressing a critical limitation of standard MLLMs: the degradation of reasoning capabilities due to the integration of multiple modalities.
\item \textbf{Cross-Modal Reasoning Adaptation:} Building on the foundational reasoning capabilities established in the first phase, this phase employs multimodal question-answer pairs augmented with caption information to progressively transfer these abilities to the multimodal domain.
\item \textbf{Multimodal Reasoning Enhancement:}
To address multimodal questions lacking comprehensive image descriptions in real-world scenarios, we further train the model using multimodal question-answer pairs, building on the foundation established in the second phase. By removing dependence on textual captions, this phase compels the model to directly interpret and reason from raw visual inputs, thereby mitigating linguistic biases and promoting robust multimodal inference.

\end{itemize}
\begin{figure}[t]
    \centering
    \includegraphics[width=0.8\linewidth]{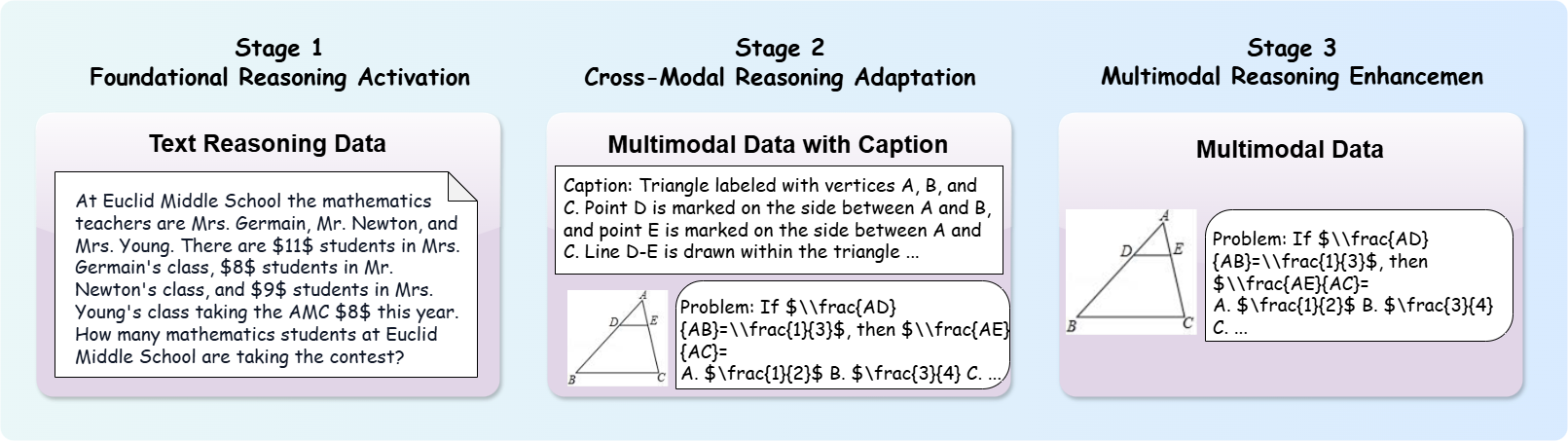}
    \caption{Utilization of Data Types Across Different Training Stages in the Infi-MMR Framework}
    \label{fig:trainingdetail}
    \vspace{-0.6cm}
\end{figure}

The data types utilized in each phase are illustrated in Figure \ref{fig:trainingdetail}. Together, our Infi-MMR framework establishes a robust pathway for restoring and enhancing reasoning capabilities in multimodal reasoning scenarios. We evaluate the efficacy of \textbf{Infi-MMR-3B}, trained using the Infi-MMR framework, on a comprehensive suite of challenging benchmarks designed to assess core mathematical reasoning abilities. The experimental results not only validate the effectiveness of our progressive training framework but also confirm the successful transfer of its reasoning capabilities to the multimodal domain. Generally, our main contributions are threefold:
\begin{itemize}[leftmargin=*, nosep]
    \item \textcolor{black}{We introduce \textbf{Infi-MMR}, a \textcolor{black}{curriculum-based} training framework comprising three phases enabling the model to build robust foundational reasoning and gradually integrating and enhancing multi-modal reasoning capabilities.}
    \item \textcolor{black}{We introduce caption-augmented multimodal data as a critical bridge to facilitate the transfer of the reasoning abilities from the textual domain to the multimodal domain, enhancing the model’s capacity for robust cross-modal inference. This dataset will be open-sourced in the future to support further exploration and advancements in multimodal reasoning.}
    \item We develop \textbf{Infi-MMR-3B}, a reasoning MSLM trained via our framework, which achieves superior results across multiple multimodal reasoning benchmarks, including MathVerse (43.68\%), MathVision (27.04\%), OlympiadBench (21.33\%), etc., demonstrating its effectiveness.

\end{itemize}

\section{Related Work}
\subsection{Multimodal Large Language Model Reasoning}
Multimodal Large Language Models (MLLMs) bridge visual perception and linguistic reasoning through architectures like Flamingo \cite{alayrac2022flamingo} and LLaVA \cite{liu2023visual}, enabling complex cross-modal tasks such as visual question answering. Current methods enhance MLLM reasoning capabilities primarily through supervised fine-tuning with high-quality multimodal Chain-of-Thought (CoT) data generated by advanced models \cite{zhang2024improve}. While effective, this approach inherits limitations in scalability due to its dependence on pre-curated reasoning traces. Models trained on fixed reasoning traces struggle to adapt to unseen domains beyond their pre-defined reasoning templates. In contrast, our work proposes a curriculum-based reinforcement learning framework that progressively unlocks multimodal reasoning.

\vspace{-0.2cm}
\subsection{Reinforcement Learning in MLLMs}
The initial deployment of reinforcement learning (RL) in LLMs primarily centered on Reinforcement Learning from Human Feedback (RLHF) \cite{ouyang2022training}. Recent advances, exemplified by DeepSeek-R1 \cite{guo2025deepseek}, have revealed RL’s capacity to directly enhance LLMs' reasoning performance. In multimodal settings, an emerging paradigm \cite{huang2025vision, yang2025r1} integrates MLLMs with DeepSeek-R1 to produce multimodal CoT data for cold-start initialization. After cold-start, the model's reasoning abilities are refined via RL training. However, this approach imposes computational overhead, as the generation of vision-grounded reasoning traces necessitates MLLMs to create comprehensive image descriptions for subsequent CoT derivation. Liu et al. \cite{liu2025x} leverage text-only SFT reasoning data to enhance MLLMs' reasoning abilities in the initial phase. Similar to \cite{peng2025lmm}, we utilize high-quality textual reasoning data to stimulate the model’s logical reasoning faculties, reducing dependency on multimodal CoT data. Additionally, instead of directly using multimodal data, we employ caption-augmented multimodal data to progressively transfer reasoning capabilities to multimodal tasks.

\vspace{-0.2cm}
\subsection{Curriculum Learning}
In curriculum learning (CL), models are exposed to data in a structured, progressive manner, starting with simpler examples and gradually increasing complexity \cite{wang2021survey}. CL has been shown to improve the model performance and accelerate the training process \cite{platanios2019competence}. Drawing inspiration from the core idea of CL, our Infi-MMR framework first activates core reasoning capabilities using text-only data. Next, the framework transfers these skills to multimodal tasks using caption-augmented data. Finally, it employs caption-free data, forcing the model to adapt to authentic multimodal challenges.

\section{Prilimary}
As MLLMs advance and integrate increasingly diverse data types, the demand to expand their multimodal capabilities intensifies. To strengthen these capabilities, many researchers have turned to Reinforcement Learning from Human Feedback (RLHF), a method designed to align model outputs with human preferences and expectations.

\subsection{Reinforcement Learning for MLLMs}
RLHF often employs Proximal Policy Optimization (PPO) \cite{schulman2017proximalppo} as a key algorithm to optimize policies during training. PPO generally involves four models:
\textbf{(1) Policy Model} generates responses to incoming pictures and questions, which guide the model's decisions.
\textbf{(2) Critic Model} estimates the expected return, which will be used as an intermediate value to calculate advantage, from a given state under the current policy. 
\textbf{(3) Reward Model} generates reward signals based on human feedback to guide the policy learning process. 
\textbf{(4) Reference Model} computes the probability ratio between the current and old policies, ensuring that updates to the policy are constrained to prevent large, destabilizing changes. However, incorporating a reward model substantially increases the computational complexity of the training process.

To effectively reduce training costs and enhance training stability, we adopt the Group Relative Policy Optimization (GRPO) algorithm \cite{guo2025deepseek} during the reinforcement learning phase. In GRPO, the advantage is computed by generating multiple responses to the same visual input, eliminating the reliance on a critic model.

Assuming we have a pre-trained MLLM and denote it as a policy model $\pi_{\theta}$. Given a multimodal question $q$, consisting of a textual task instruction and one or more images, i.e. $q=\{x,\mathcal{I}\}$, the policy model $\pi_{\theta_{old}}$ (before current update) generates $G$ candidate outputs $\{o_i\}_{i=1}^{G}$. For each output $o_i$, we use a rule-based reward function $R(o,q)$ to evaluate the quality of the output. Based on these rewards $r_i$, we calculate the group-relative advantage $A_i$ as follows:
\begin{equation}
    A_i=\frac{r_i-mean(\{r_1,r_2,\dots,r_G\})}{std(\{r_1,r_2,\dots,r_G\})},
    \label{grpo:advantage}
\end{equation}
where $mean(\cdot)$ denotes the average and $std(\cdot)$ represents the standard deviation.

Based on the above, to obtain a better policy model $\pi_{\theta}$, we maximize the $ \mathcal{J}_{\text{GRPO}}(\theta)$ objective function 

\begin{align}
& \mathcal{J}_{\text{GRPO}}(\theta) = \mathbb{E}_{[q \sim P(Q), \{o_i\}_{i=1}^G \sim \pi_{\theta_{\text{old}}}(O|q)]}  
\frac{1}{G} \sum_{i=1}^{G} \frac{1}{|o_i|} \nonumber \\
& \sum_{t=1}^{|o_i|} \left\{ \min \left[ \frac{\pi_{\theta}(o_{i}|q)}{\pi_{\theta_{\text{old}}}(o_{i}|q)} {A}_{i}, \text{clip} \left( \frac{\pi_{\theta}(o_{i}|q)}{\pi_{\theta_{\text{old}}}(o_{i}|q)}, 1-\epsilon, 1+\epsilon \right) A_{i} \right] - 
 \beta D_{KL} \left[ \pi_{\theta} \| \pi_{\text{ref}} \right] \right\},
\end{align}
where $o_i$ is the $i_{th}$ generated output. The additional Kullback–Leibler term $D_{KL} \left[ \pi_{\theta} \| \pi_{\text{ref}} \right]$ is applied to penalize divergence from a reference model $\pi_{ref}$, helping prevent catastrophic forgetting. $\epsilon, \beta \in \mathbb{R} \geq 0$ control the regularization strengths to stabilize the training process.

\subsection{Design of the Reward Function}
The design of the reward function $R(o,q)$ is crucial for guiding the policy model to learn a reasoning trajectory. Our total reward $R_{total}$ integrates assessments of both output format correctness and accuracy:
\begin{equation}
    R_{total}(o, q) = w_f \cdot R_{format}(o, q) + w_a \cdot R_{acc}(o, q)
\end{equation}
where $R_{format}(o, q)$ is the reward for output format correctness and $R_{acc}(o, q)$ is the reward for accuracy, given an output $o$ for an input query $q$. The coefficients $w_f$ and $w_a$ are non-negative hyperparameters that weight the relative importance of these two components, satisfying $w_f + w_a = 1$.

The format reward, $R_{format}(o, q)$, assesses whether the model's output $o$ adheres to predefined structural requirements. Specifically, it verifies two primary aspects:
\begin{itemize}[leftmargin=*, nosep]
    \item \textbf{Thinking Process Format:} It checks if the model correctly presents its reasoning process using the specified format. For instance, the model might be instructed to encapsulate its reasoning within specific tags. An example of a system prompt used during RL training and inference to enforce such a format is shown below:
    
    \begin{tcolorbox}[
        colback=blue!6!white,
        colframe=black,
        colbacktitle=black,
        coltitle=white,
        fonttitle=\bfseries\sffamily,
        title=System Prompt for RL Training and Inference,
        sharp corners,
        boxrule=1pt,
    ]
    You FIRST think about the reasoning process as an internal monologue and then provide the final answer.
    
    The reasoning process MUST BE enclosed within \texttt{<think> </think>} tags.
    \end{tcolorbox}

    \item \textbf{Final Answer Provision:} It confirms whether a final answer is explicitly provided by the model, particularly when the instructions associated with the query $q$ require such an output.
\end{itemize}
$R_{format}(o, q)$ is a binary reward, yielding a value of 1 if all specified format criteria are met, and 0 otherwise.

The accuracy reward, $R_{acc}(o, q)$, quantifies the accuracy of the final result in the model's output $o$ with respect to the ground truth for query $q$. A critical aspect of our reward design is that $R_{acc}(o, q)$ is computed \textit{only if} the output is structurally sound, i.e., when $R_{format}(o, q) = 1$. This staged approach encourages the model to first learn to generate well-formed outputs before focusing on the accuracy of the result. If $R_{format}(o, q) = 0$, then $R_{acc}(o, q)$ is effectively zero. The methodology for calculating $R_{acc}(o, q)$ when $R_{format}(o, q) = 1$ is tailored to the nature of the task, which we categorize based on the ground truth answer format:
\begin{itemize}[leftmargin=*, nosep]
    \item \textbf{Mathematical Tasks:} For tasks involving mathematical expressions or numerical answers, $R_{acc}(o, q)$ is determined by a specialized verification function, denoted \texttt{math\_verify}$(o_{ans}, gt_{ans})$. This function evaluates the extracted answer from the output $o$, denoted $o_{ans}$, against the ground truth answer $gt_{ans}$. The \texttt{math\_verify} function is designed to handle nuances of mathematical evaluation, potentially allowing for symbolic equivalence or specified numerical tolerances. A successful verification yields a reward of 1, otherwise 0.
    \item \textbf{String-based Tasks:} For tasks where the expected answer is textual, $R_{acc}(o, q)$ is determined by comparing the model's generated answer string with the ground truth string after a normalization process. Normalization procedures typically include operations such as conversion to lowercase and the removal of leading/trailing whitespace and redundant internal spaces. An exact match between the normalized $o_{ans}$ and normalized $gt_{ans}$ yields a reward of 1; otherwise, the reward is 0.
    \item \textbf{Multiple-Choice Questions:} For tasks requiring the selection of an option from a predefined set, $R_{acc}(o, q)$ is determined by a direct comparison of the model's selected choice ($o_{ans}$) with the correct ground truth option ($gt_{ans}$). A match yields a reward of 1, and a mismatch results in a reward of 0.
\end{itemize}

\begin{figure}[t]
    \vspace{-0.4cm}
    \centering
    \includegraphics[width=1.0\linewidth]{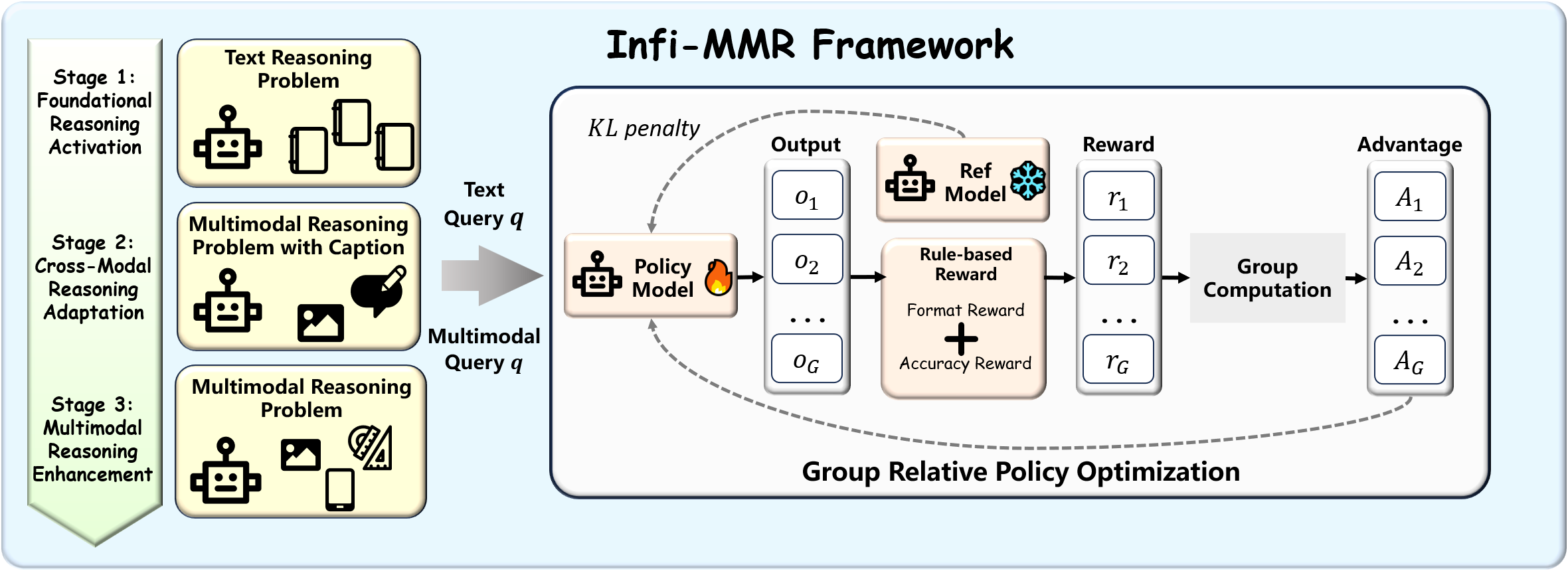}
    \caption{The Overall Framework of Infi-MMR.}
    \label{fig:framework}
    \vspace{-0.4cm}
\end{figure}

\section{Methodology}\label{sec:methodology}
To address the aforementioned challenges and enhance the reasoning capabilities of MSLMs, we propose a novel framework, Infi-MMR. As illustrated in Figure \ref{fig:framework}, Infi-MMR employs a curriculum of three distinct rule-based reinforcement learning phases. The the first phase, Foundational Reasoning Activation (FRA), leverages text-only mathematical reasoning datasets to activate and fortify the core reasoning capabilities of MSLMs. The second phase, Cross-Modal Reasoning Adaptation (CMRA), facilitates the transfer of these reasoning abilities to multimodal contexts through the use of caption-augmented multimodal data. The third phase, Multimodal Reasoning Enhancement (MRE), utilizes caption-free multimodal data to eliminate linguistic biases and strengthen pure cross-modal reasoning, thereby unlocking the full reasoning potential of MSLMs.
 
\subsection{Phase 1. Foundational Reasoning Activation}
To activate and enhance the foundational reasoning ability of the base MSLMs, limited by the lack of high-quality multimodal reasoning data, we first utilize large-scale and high-quality verifiable text-only data for rule-based RL in this initial phase. This approach harnesses an extensive range of text-based reasoning questions, which are inherently more difficult and require sophisticated reasoning processes than many current multimodal reasoning tasks. By engaging with these comprehensive textual reasoning exercises, we aim to cultivate robust foundational reasoning skills within the model, which can subsequently be adapted to multimodal scenarios.

\subsection{Phase 2. Cross-Modal Reasoning Adaptation}
After assessing the robustness and adaptability of the model’s foundational reasoning skills, we progressively transfer these capabilities into the multimodal domain.

To achieve this objective, we employ caption-augmented multimodal data to facilitate the transfer of reasoning skills. Captions serve as a crucial bridge, connecting text-based reasoning with multimodal comprehension by providing contextual descriptions that link visual inputs to structured linguistic frameworks.

To efficiently and accurately generate image captions, we utilized Omnicaptioner~\cite{omnicaptioner}, a framework designed for generating captions across various visual domains at different levels of granularity. For diverse image types, we first employed Qwen2.5-VL-7B \cite{Qwen2.5-VL} with a specific instruction, which is presented in the Appendix \ref{appendix:A}, to classify them into distinct categories.

For each image category, we applied the primary system prompt in Omnicaptioner to generate a concise caption. Subsequently, by augmenting the problem with generated captions, we utilize RL to progressively transfer the model's foundational reasoning skills to the multimodal domain. Examples of generated captions are shown in the Appendix \ref{appendix:B}

\subsection{Phase 3. Multimodal Reasoning Enhancement}
After initially transferring reasoning capabilities to the multimodal domain using caption-augmented data, the model must eliminate its dependence on textual information and enhance its capacity for pure vision reasoning.

To ensure the agent strengthens its math-related reasoning skills while preserving its general multimodal understanding and reasoning capabilities, we leverage a diverse collection of high-quality, verifiable multimodal reasoning datasets. These datasets span a broad range of topics and difficulty levels—from grade school problems to advanced STEM subjects—and incorporate visual elements such as charts, diagrams, and spatial relationships. Following the CMRA phase, the model undergoes training across varied visual contexts and reasoning tasks simultaneously.

The transition from caption-augmented to raw multimodal data is a deliberate design choice to bolster the model’s capabilities. By removing textual captions, the model is compelled to interpret and reason solely based on visual inputs, without supplementary linguistic cues. This shift enhances the model’s cross-modal reasoning proficiency, enabling it to independently process and integrate information across modalities. Consequently, the model becomes more adaptable and effective in addressing a wide array of multimodal tasks where textual support may be unavailable.

\section{Experiments}\label{sec:experiments}
In this section, we elaborate on the experimental setup employed to train and evaluate our proposed \textbf{Infi-MMR-3B} model, which is based on Qwen2.5-VL-3B-Instruct~\cite{Qwen2.5-VL}. We provide a detailed description of the implementation details, the evaluation benchmarks, and a comprehensive analysis of the results compared to state-of-the-art methods. Additionally, we also analyzed the effects of each training phase.

\subsection{Experimental Setup}\label{sec:experiment_setup}
\textbf{Implementation Details.} Our model, \textbf{Infi-MMR-3B}, is built upon Qwen2.5-VL-3B-Instruct and trained using the proposed \textbf{Infi-MMR} Framework, which consists of three main phases.
For the RL reward function $R_\text{total} = w_f \cdot R_\text{format} + w_a \cdot R_\text{acc}$, we set the weights $w_f = 0.1$ and $w_a = 0.9$. \textcolor{black}{Within the mathematical accuracy reward $R_\text{acc\_math}$ and multiple-choice rewards ($R_\text{choice}$), the weights are $w_t = 0.2$ for type matching and $w_p = 0.8$ for exact parameter matching.} All experiments were conducted using 16 NVIDIA H800 GPUs. For each phase, we used a learning rate of 1.0e-6, a batch size of 256 for training updates, a rollout batch size of 256, and generated 16 rollouts per sample during policy exploration.

\textbf{Training Data.} To establish robust multimodal reasoning capabilities, we initially train \textbf{Infi-MMR-3B} on DeepScaleR \cite{deepscaler2025}, a high-quality textual reasoning dataset comprising 39,000 verifiable mathematics problem-answer pairs. In the second and third phases, we leverage ViRL39k \cite{vl-rethinker}, a dataset containing 39,000 verifiable question-answer pairs involving charts, tables, spatial relationships, and image understanding. Specifically, during the Cross-Modal Reasoning Adaptation phase (Phase 2), we classify each image and employ Omnicaptioner \cite{omnicaptioner} to generate a concise caption, facilitating the integration of visual and textual reasoning. 

\textbf{Decontamination.} We implement a two-stage decontamination process to ensure a fair and robust evaluation of multimodal language model performance. In the first stage, inspired by the approach employed in Light-R1 \cite{wen2025light}, we apply a 32-gram based text deduplication and execute exact matching after removing numerical information to account for samples that differ only in numerical content. In the second stage, we extract multimodal embeddings from both the training and test sets with gme-Qwen2-VL-2B-Instruct model \cite{zhang2024gme}. Samples exceeding a similarity threshold of 0.95 are removed. This approach helps mitigate data leakage and ensures that the evaluation remains unbiased and reflective of true generalization capabilities.

\subsection{Evaluation Benchmarks}

To comprehensively evaluate \textbf{Infi-MMR-3B}, we utilize several key benchmarks targeting different facets of reasoning capabilities:

\textbf{MATH500~\cite{lightman2023let}:} This benchmark comprises 500 mathematical problems spanning algebra, geometry, probability, and other topics, designed to assess the model's mathematical reasoning capabilities in textual reasoning tasks. \\
\textbf{MathVerse~\cite{zhang2024mathverse} testmini \& MathVision~\cite{wang2024mathvision} test \& OlympiadBench~\cite{he2024olympiadbench}:} These benchmarks evaluate the model's proficiency in performing reasoning-dominant tasks within the multimodal domain. MathVerse, with its diverse question types, assesses the extent to which MLLMs can comprehend visual diagrams. MathVision offers a comprehensive and varied set of problems to test reasoning breadth. OlympiadBench, featuring Olympiad-level questions, gauges the model’s capacity to tackle complex, high-difficulty problems. \\
\textbf{MathVista~\cite{lumathvista} testmini:} This benchmark presents a curated set of reasoning problems designed to evaluate the model’s general multimodal capabilities.

\begin{table*}[t]
    \centering
    \small
    \caption{Performance comparison of different MLLMs across various reasoning-related benchmarks. Results colored in \textcolor{red}{red} represent the best performance, and those \underline{underlined} indicate the suboptimal performance. }
    \label{table:overallresults}
    \resizebox{0.95\linewidth}{!}
{
    \begin{tabular}{lcccccc}
    \toprule
    \multirow{3}{*}{\textbf{Model}} & \multicolumn{5}{c}{\textbf{Accuracy (\%)}} \\
    \cmidrule(lr){2-6}
    & \multicolumn{1}{c}{Text-Only Reasoning} & \multicolumn{3}{c}{Multimodal Reasoning} & \multicolumn{1}{c}{Multimodal General} \\
    \cmidrule(lr){2-2} \cmidrule(lr){3-5} \cmidrule(lr){6-6}
    & MATH500 & MathVerse & MathVision & OlympiadBench & MathVista & \\
    \midrule
    \rowcolor{gray!15}
    \multicolumn{6}{l}{\textit{Proprietary Models}} \\
    GPT-4o~\citep{hurst2024gpt}  & - & 39.4 & 30.4 & - & 63.8 \\
    \midrule
    \rowcolor{gray!15}
    \multicolumn{6}{l}{\textit{Base Model Qwen2-VL-7B}} \\
    Qwen2-VL-7B~\citep{wang2024qwen2} & - &  31.9 & 18.8 & - & 58.2  \\
    Mulberry~\citep{yao2024mulberry} & - &  39.5 & 23.4 & - &  62.1 \\
    \midrule
    \rowcolor{gray!15}
    \multicolumn{6}{l}{\textit{Based Model InternVL2-8B}} \\
    InternVL2-8B~\citep{chen2024expanding} & - & - & 20.4 & - & 58.3 \\
    InternVL2-8B-MPO~\citep{chen2024expanding} & - & - & 25.7 & - & 67.0  \\
    \midrule
    \rowcolor{gray!15}
    \multicolumn{6}{l}{\textit{Based Model InternVL2.5-8B}} \\
    InternVL2.5-8B~\citep{chen2024expanding} & - & 39.5 & 19.7 & 8.0 & 64.4 \\
    MM-Eureka-8B~\citep{meng2025mm} & - & 40.4 & 22.2 & 8.6 & \underline{67.1}  \\
    \midrule
    \rowcolor{gray!15}
    \multicolumn{6}{l}{\textit{Based Model Qwen2.5VL-3B}} \\
    Qwen2.5-VL-3B~\citep{Qwen2.5-VL} & 63.40 & 33.20 & 21.25 & 11.33 & 63.40\\
    FRE-TEXT-3B~\citep{peng2025lmm} & 65.4 & 38.83 & 25.76 & 15.62 & 61.4  \\
    MGT-PerceReason-3B~\citep{peng2025lmm}  & 63.80 & 41.55 & \underline{26.35} & 15.62 & 63.20 \\
    FAST-3B~\citep{xiao2025fast}  & - &  \underline{43.0} &  26.8 & 14.67 & 66.2  \\
    \rowcolor{gray!15}
    \multicolumn{6}{l}{\textit{Ours}} \\
    Infi-MMR\_FRA & \textcolor{red}{68.8} & 40.8 & 23.91 & \underline{19.33} & 62.9  \\
    Infi-MMR\_CMRA & \underline{65.65} & 42.84 & 26.34 & \underline{19.33} & 63.5 \\
    \textbf{Infi-MMR-3B} & 65.5 &  \textcolor{red}{43.68} &  \textcolor{red}{27.04} & \textcolor{red}{21.33} & \textcolor{red}{67.2}  \\
    \bottomrule
    \end{tabular}
}
\end{table*}

\subsection{Main Results}
We compare \textbf{Infi-MMR-3B} against a range of state-of-the-art open-source and proprietary reasoning-focused MLLMs, the results are summarized in Table \ref{table:overallresults}, where Infi-MMR\_FRA and Infi-MMR\_CMRA are reasoning-enhanced models via the FRA phase and the CMRA phase, respectively.

On the MATH500 benchmark, our Infi-MMR series achieved the highest scores among all compared models. Notably, Infi-MMR\_FRA attained the highest accuracy of $68.8\%$, representing a $5.4\%$ improvement over the base model. In the multimodal domain, all Infi-MMR series models demonstrate distinct improvements in reasoning strength compared to Qwen2.5-VL-3B. In particular, Infi-MMR-3B achieves state-of-the-art performance, surpassing all compared models, including MLLMs built on the same base model and those with larger parameter counts. Across the MathVerse, MathVision, and OlympiadBench benchmarks, Infi-MMR-3B recorded accuracies of $43.68\%$, $27.04\%$, and $21.33\%$, respectively, showcasing robust reasoning capabilities on diverse multimodal mathematical reasoning problems of varying types and difficulties.

Moreover, the progressive increase in performance from Infi-MMR\_FRA to Infi-MMR-3B across multimodal benchmarks reflects the efficacy of the phased approach. The initial text-only training establishes a strong reasoning base, the caption-augmented phase bridges to multimodal contexts with moderate success, and the caption-free phase optimizes performance by enhancing multimodal reasoning capability. 
Additionally, on the multimodal general benchmark MathVista, Infi-MMR\_FRA exhibits a performance decline compared to Qwen2.5-VL-3B. In contrast, Infi-MMR\_CMRA and Infi-MMR-3B achieve improvements of $0.1\%$ and $3.8\%$, respectively, highlighting the importance of interpreting visual inputs to mitigate linguistic biases effectively.

\subsection{Ablation Study}
In this subsection, we aim to address the rationale behind adopting a three-phase training framework by answering the following research questions:

\begin{itemize}[leftmargin=*,nosep]
    \item \textbf{RQ1:} Why is direct application of multimodal RL unsuitable for the initial phase?
    \item \textbf{RQ2:} Why is it effective to use caption-augmentation data? 
\end{itemize}

\begin{figure}
    \centering
    \includegraphics[width=0.95\linewidth]{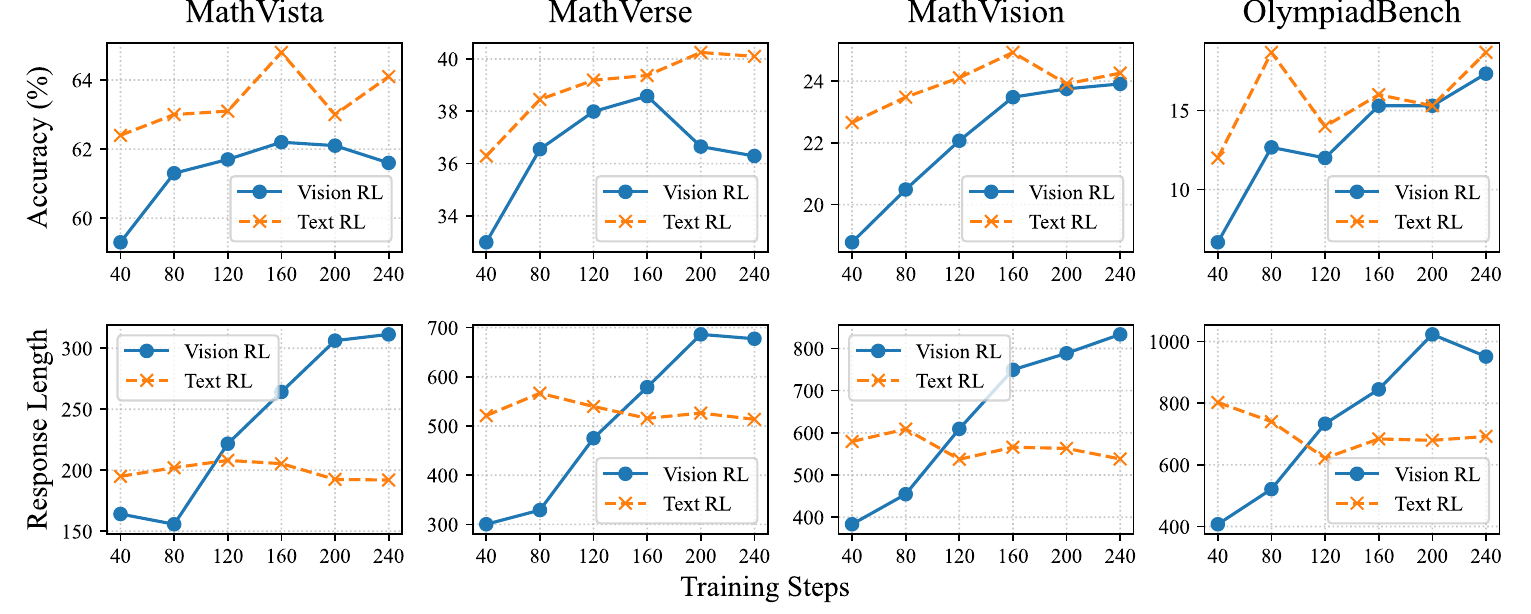}
    \caption{Analysis of Different Modality Data for Initial Training. Text RL and Vision RL represent the types of data utilized in the initial RL phase.}
    \label{fig:ablation_1st}
    \vspace{-0.4cm}
\end{figure}

\subsubsection{Analysis of Different Modality Data for Initial Training (RQ1)}
To show the influence of different modalities of data in the initial training phase, we conducted experiments using text-only data and multimodal data, respectively, while maintaining consistent hyperparameter settings across both experiments. The performance on multimodal benchmarks and the average response token length are illustrated in \textcolor{black}{Figure \ref{fig:ablation_1st}}, where tokens are counted with Qwen2.5-VL’s tokenizer. 

Performance across various training steps demonstrates that MSLMs trained with text-only data in the initial phase consistently outperform those trained with multimodal data on multimodal reasoning benchmarks. This finding underscores the effectiveness of using text-only data to establish stronger foundational reasoning capabilities in MSLM, while maintaining competitive performance on multimodal tasks during the initial training phase.

Additionally, we analyzed the average response length across training steps and identified distinct trends based on the training data modality. With text-only data, the average token count of responses initially rises, then declines, and stabilizes, reflecting a controlled adaptation of the model’s reasoning process. Conversely, training with multimodal data leads to a steadily increasing response length, eventually exceeding the maximum observed with text-only training, yet yielding limited performance improvements. Moreover, multimodal training introduces instability, as demonstrated by performance declines on the MathVerse and MathVision despite increased response lengths, suggesting the generation of longer yet less meaningful outputs. This finding underscores the importance of initiating multimodal reinforcement learning with text-only data to ensure stable and effective reasoning development.

\begin{table*}[t]
    \centering
    \small
    \caption{Performance comparison of different data types used in the second stage, continuing training from Infi-MMR\_FRP, on multimodal reasoning benchmarks. Results colored in \textcolor{Red}{Red} represent the best performance.}
    \label{table:ablationstudy_2nd_stage}
    \resizebox{0.75\linewidth}{!}
{
    \begin{tabular}{lcccc}
    \toprule
    \multirow{2}{*}{\textbf{Model}} & \multicolumn{4}{c}{\textbf{Accuracy (\%)}} \\
    \cmidrule(lr){2-5}
    & MathVerse & MathVision & OlympiadBench & MathVista  \\
    \midrule
    Infi-MMR\_FRA  &  40.8 & 23.91 & 19.33 & 62.9 \\
    \midrule
    \rowcolor{gray!15}
    \multicolumn{5}{l}{\textit{Rule-Based RL on Caption-Free Multimodal Dataset}} \\
    Infi-MMR\_CapFre & 41.94 & 25.88 & 18.67  & \textcolor{Red}{63.9} \\
    \midrule
    \rowcolor{gray!15}
    \multicolumn{5}{l}{\textit{Rule-Based RL on Caption-Augmented Multimodal Dataset}} \\
    Infi-MMR\_CMRA & \textcolor{Red}{42.84} & \textcolor{Red}{26.34} & \textcolor{Red}{19.33} & 63.5  \\
    \bottomrule
    \end{tabular}
}
    \vspace{-0.2cm}
\end{table*}

\subsubsection{Analysis of the Effectiveness of Caption-Augmentation Data (RQ2)}
To illustrate the impact of caption-augmentation data in our Infi-MMR framework, we additionally continue the rule-based RL training on the Infi-MMR\_FRA model with the original ViRL39K dataset. The results on multimodal reasoning benchmarks are illustrated in Table \ref{table:ablationstudy_2nd_stage}. 
It was noticed that the caption-free rule-based RL (Infi-MMR\_CapFre) achieves a higher accuracy of 63.9\%, compared to the caption-augmented approach (Infi-MMR\_CMRA). 
This superior performance arises from the model’s ability to directly acquire visual recognition and interpretation skills from caption-free data, a capability absent in the caption-augmented approach due to its dependence on textual descriptions. However, this enhanced visual proficiency comes at the expense of a moderated improvement in multimodal reasoning capabilities. On reasoning-specific benchmarks, the caption-free method yields suboptimal results and even exhibits performance degradation on OlympiadBench, indicating a trade-off. This suggests that while caption-free data boosts general multimodal performance through improved visual learning, it may constrain the depth of reasoning development by directly transferring reasoning abilities to the multimodal domain without sufficient guidance.


\section{Conclusion}\label{sec:conclusion}
We present \textbf{Infi-MMR-3B}, a multimodal small language model focused on deliberative reasoning capabilities. Through the Infi-MMR framework, our approach systematically restores and enhances reasoning capabilities in MLLMs via three rule-based reinforcement learning phases: (1) \textbf{Foundational Reasoning Activation}, which restores and builds foundational reasoning capabilities using text-only reasoning data; (2) \textbf{Cross-Modal Reasoning Adaptation}, which utilizes caption-augmented multimodal data to progressively transfer reasoning abilities to multimodal tasks; and (3) \textbf{Multimodal Reasoning Enhancement}, which eliminates reliance on textual captions, thereby mitigating linguistic biases and promoting robust multimodal inference. Empirical results across diverse benchmarks demonstrate that Infi-MMR-3B achieves state-of-the-art accuracy compared to MLLMs with the same base model, even surpassing some MLLMs with more parameters.
Despite these promising outcomes, this study has limitations. Notably, the quality of generated captions for the Cross-Modal Reasoning Adaptation phase was not a primary research focus, and its precise impact on the final results warrants further investigation.
In addition, our method has no negative social impact.

\clearpage
\newpage
\section*{References}
\bibliographystyle{IEEEbib}
\setstretch{1.15}
\bibliography{infimmr}

\newpage
\appendix
\setcounter{section}{0}
\section{Examples of Generated Caption}
\label{appendix:B}
\vspace{-0.4cm}
\begin{figure}[h]
    \centering
    \includegraphics[width=1.0\linewidth]{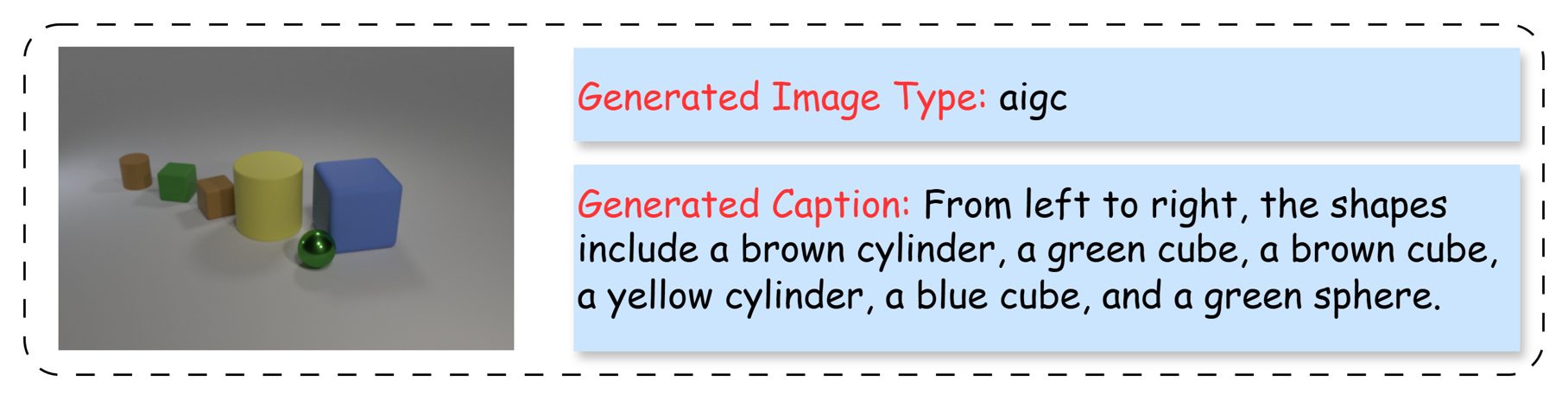}
    \vspace{-0.2cm}
    \caption{Generated Caption With aigc Type.}
    \label{fig:enter-label}
    \vspace{-0.4cm}
\end{figure}

\begin{figure}[h]
    \centering
    \includegraphics[width=1.0\linewidth]{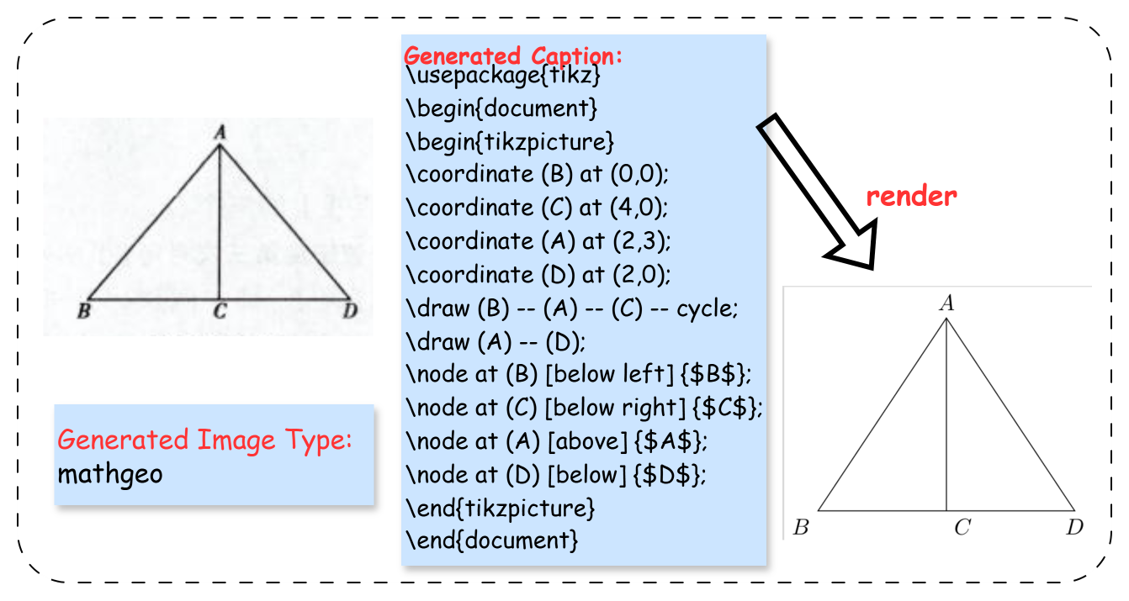}
    \vspace{-0.2cm}
    \caption{Generated Caption With mathgeo Type.}
    \label{fig:enter-label}
    \vspace{-0.4cm}
\end{figure}

\begin{figure}[h]
    \centering
    \includegraphics[width=1.0\linewidth]{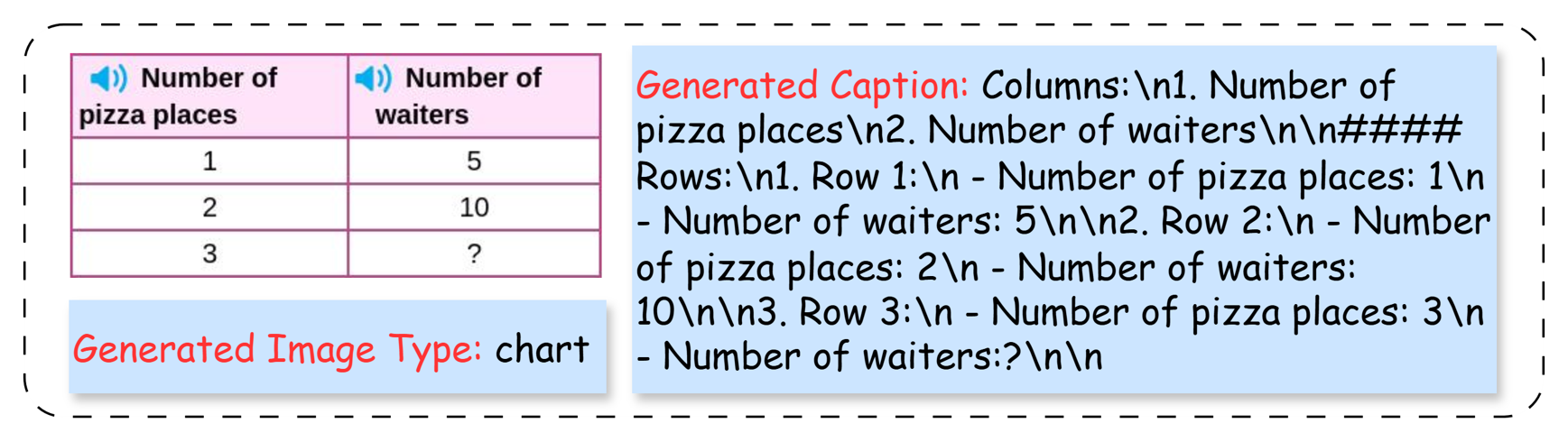}
    \vspace{-0.2cm}
    \caption{Generated Caption With table Type.}
    \label{fig:enter-label}
    \vspace{-0.4cm}
\end{figure}

\begin{figure}[h]
    \centering
    \includegraphics[width=1.0\linewidth]{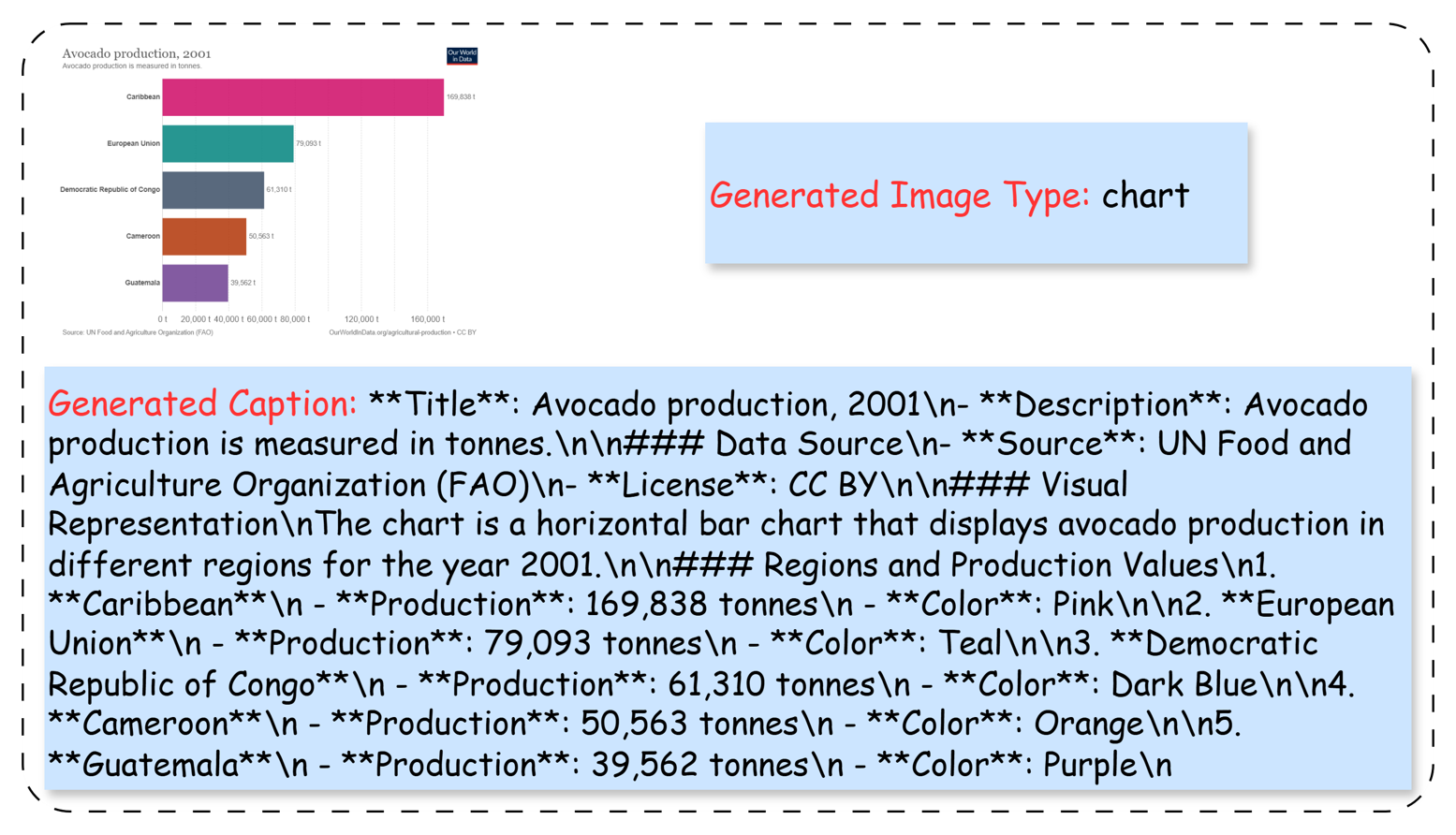}
    \vspace{-0.2cm}
    \caption{Generated Caption With chart Type.}
    \label{fig:enter-label}
\end{figure}

\section{Instruction for Clarifying Images}\label{appendix:A}

\begin{tcolorbox}[
    colback=blue!6!white,
    colframe=black,
    colbacktitle=black,
    coltitle=white,
    fonttitle=\bfseries\sffamily,
    title=Instruction for Clarifying Images,
    sharp corners,
    boxrule=1pt
]
Classify the image into exactly one of the following categories: \\

For structured data (can be written in Markdown or Latex): \textbf{mathgeo} (Euclidean geometric shapes or mathematical related diagrams), \textbf{chart} (scatter plots, bar charts, line graphs), \textbf{table} (data tables),\\
For unstructured data, flowcharts, mixed types, complex scenes, or data not belonging to the above categories: \textbf{aigc}, \\

Respond with the category NAME ONLY (e.g., 'mathgeo'). Do not include any additional text, explanations, or symbols.
\end{tcolorbox}

\end{document}